\documentclass[11pt]{article}

\usepackage[preprint]{acl}

\usepackage{times}
\usepackage{latexsym}
\usepackage{amsmath,amssymb}
\usepackage{ulem}

\usepackage[T1]{fontenc}

\usepackage[utf8]{inputenc}

\usepackage{microtype}

\usepackage{inconsolata}
\usepackage{enumitem}

\usepackage{graphicx}
\usepackage[most]{tcolorbox}
\usepackage{subcaption}
\usepackage{wrapfig}
\usepackage{booktabs}
\usepackage[table]{xcolor}
\usepackage{listings}

\newcommand{\mypara}[1]{\vspace{0.5em}\noindent\textbf{#1}}

\definecolor{mygreen}{HTML}{00A64F}
\definecolor{HeaderGreen}{RGB}{47, 95, 71}     %
\definecolor{BodyGreen}{RGB}{238, 245, 238}    %

\newtcolorbox{takeawaybox}[1]{
    enhanced,
    colback=BodyGreen,
    colframe=HeaderGreen,
    colbacktitle=HeaderGreen,
    coltitle=white,            %
    fonttitle=\bfseries,       %
    arc=6pt,                   %
    titlerule=0pt,             %
    title={#1},                %
    left=4pt,
    right=4pt,
    top=4pt,
    bottom=4pt,
    toptitle=1pt,              %
    bottomtitle=1pt,
    boxrule=1pt                %
}

\usepackage[colorinlistoftodos]{todonotes}
\definecolor{amber}{HTML}{FFBF00}

\title{On Asymmetric Optimization of Reasoning and Perception in Vision-Language Model Post-Training}

\author{Xueqing Wu,~~Yu-Chi Lin,~~Kai-Wei Chang,~~Nanyun Peng \\
University of California, Los Angeles \\
\url{https://asymmetric-vlm-post-training.github.io}}

\begin{document}
\maketitle

\begin{abstract}
Post-training has greatly improved reasoning in frontier vision-language models, yet its gains for perception remain comparatively limited, creating a bottleneck for end-to-end visual reasoning. To investigate this gap, we introduce a \textit{controlled diagnostic framework} with two synthetic tasks that disentangle perception from reasoning. Our analysis reveals a consistent \textbf{perception-reasoning asymmetry}: post-training improves reasoning more substantially than perception, though the underlying mechanism differs across training paradigms. For supervised fine-tuning (SFT), this asymmetry stems from \textbf{token imbalance}, with perception occupying a smaller fraction of tokens in chain-of-thought supervision. Reweighting the loss   boosts end-to-end performance by up to $18.2$ points. For reinforcement learning (RL), the asymmetry instead arises from \textbf{reward coupling}, as outcome rewards correlate more strongly with reasoning than perception. Adding a perception-aware reward improves end-to-end accuracy by up to $6.0$ points; when ground-truth perception rewards are unavailable, a reliable surrogate provides useful signal, yielding gains of $2.2$ points. Beyond the controlled setting, these  strategies also improve real-world visual reasoning, with gains of up to $3.3$ points across three benchmarks. Overall, we diagnose the causes of asymmetric optimization and provide actionable guidance that benefits both synthetic and realistic settings.

\end{abstract}

\begin{figure}
    \centering
    \includegraphics[width=\linewidth]{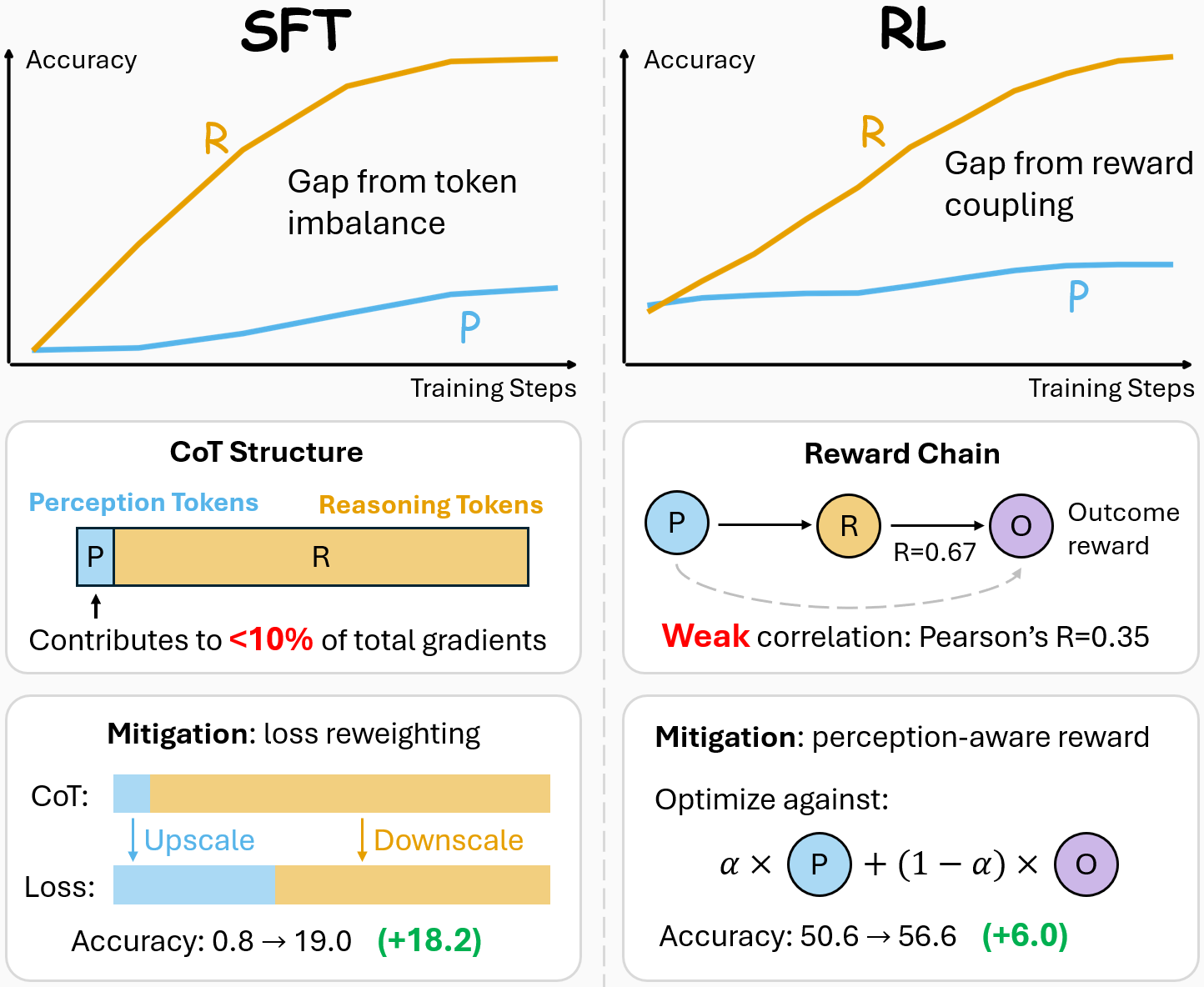}
    \vspace{-18pt}
    \caption{\small\textbf{Asymmetric optimization of perception and reasoning in VLM post-training.} Both SFT (left) and RL (right) improve reasoning ($R$) substantially more than perception ($P$), but with different causes and mitigation mechanisms.}
    \vspace{-14pt}
    \label{fig:teaser}
\end{figure}

\section{Introduction}

Post-training methods, especially supervised fine-tuning (SFT) and reinforcement learning (RL), have become central to modern vision-language models (VLMs). Frontier VLMs trained with these techniques exhibit increasingly strong reasoning capabilities \citep{deng2025openvlthinker,wang2025vl,bai2025qwen3vltechnicalreport}, achieving state-of-the-art performance across a wide range of benchmarks \citep{lumathvista,yue2024mmmu,yue2025mmmu}. However, stronger reasoning does not necessarily imply more reliable perception: recent studies suggest that post-training may yield limited gains, or even degradations, in perception \citep{liu2025more,tian2025more,yao2025rethinking}. These observations raise a central question: \textit{why does post-training disproportionately favor reasoning over perception?} Understanding the cause of this imbalance is critical, as perception errors can severely bottleneck end-to-end visual reasoning performance.

To systematically study this question, we design \textbf{controlled settings} for two complementary analyses: (1) direct evaluation of model-predicted perception and (2) isolated evaluation of reasoning via counterfactual inference conditioned on ground-truth perception. These analyses require tasks with canonical perception annotations and structured CoT format that cleanly separates perception and reasoning outputs. We instantiate the framework with graph coloring and Sudoku, using programmatically synthesized canonical perceptions and their rendered images.

Within this framework, we   experiment  with two strong open-source VLM families, Qwen3-VL \citep{bai2025qwen3vltechnicalreport} and InternVL3.5 \citep{wang2025internvl3}. We identify \textbf{a consistent asymmetric learning behavior}, where both SFT and RL improve reasoning substantially more than perception, leaving perception as the dominant bottleneck for end-to-end visual reasoning.
To explain this asymmetry, we investigate two \textbf{mechanisms}: \textit{token imbalance}, where perception  receives little optimization signal because it occupies   a smaller fraction of chain-of-thought (CoT) tokens; and \textit{reward coupling}, where RL outcome rewards align more strongly with reasoning than with perception.

Empirical results in SFT identify \textbf{token imbalance} as the primary driver: perception accounts for only $2.5\%$ of output tokens and contributes just $1.3\%$ of the final loss. A targeted loss-reweighting intervention confirms this mechanism but exposes a \textit{perception-reasoning trade-off}: though both abilities improve during training, they compete for a shared optimization budget, so upweighting the perception loss produces larger perception gains while tempering the gains in reasoning. Crucially, end-to-end accuracy peaks when perception is \textit{modestly upweighted}, improving over standard SFT by up to $13.8$ points. To avoid manual tuning, we further apply dynamic multi-task balancing methods and improve over standard SFT by up to $18.2$ points.

RL results instead suggest \textbf{reward coupling} as the primary driver: outcome rewards correlate substantially more strongly with reasoning correctness ($r\!=\!0.65$--$1.00$) than with perception correctness ($r\!=\!0.34$--$0.43$). Adding a perception reward confirms this mechanism and again exposes the \textit{perception-reasoning trade-off}; in most settings, end-to-end performance peaks at a \textit{non-zero weight} on perception reward, improving over standard RL by up to $6.0$ points. Since ground-truth perception rewards are often unavailable, we also evaluate surrogates such as teacher-model rewards. Correlation with perception accuracy proves an effective \textit{diagnostic} of perception gains, with the strongest surrogate improving performance by $2.2$ points.

Extending beyond controlled settings, we show that the identified mitigations transfer effectively to real-world visual reasoning tasks. Across three benchmarks, dynamic multi-task balancing improves standard SFT by up to $3.3$ points, while perception-aware reward augmentation provides gains of up to $1.3$ points, highlighting the broader practical value of these mitigation strategies.

Overall, our contributions are:
(1) We introduce a controlled framework for disentangling perception and reasoning in VLM post-training;
(2) We identify a consistent perception-reasoning asymmetry and trace it to distinct mechanisms: token imbalance in SFT and reward coupling in RL;
(3) We demonstrate that targeted mitigations, specifically loss reweighting for SFT and perception rewards for RL, reduce this asymmetry, improving end-to-end reasoning accuracy   by up to $18.2$ points and further transferring to real-world benchmarks.

\section{Analysis Framework}

\begin{figure*}[!t]
    \centering
    \includegraphics[width=\linewidth]{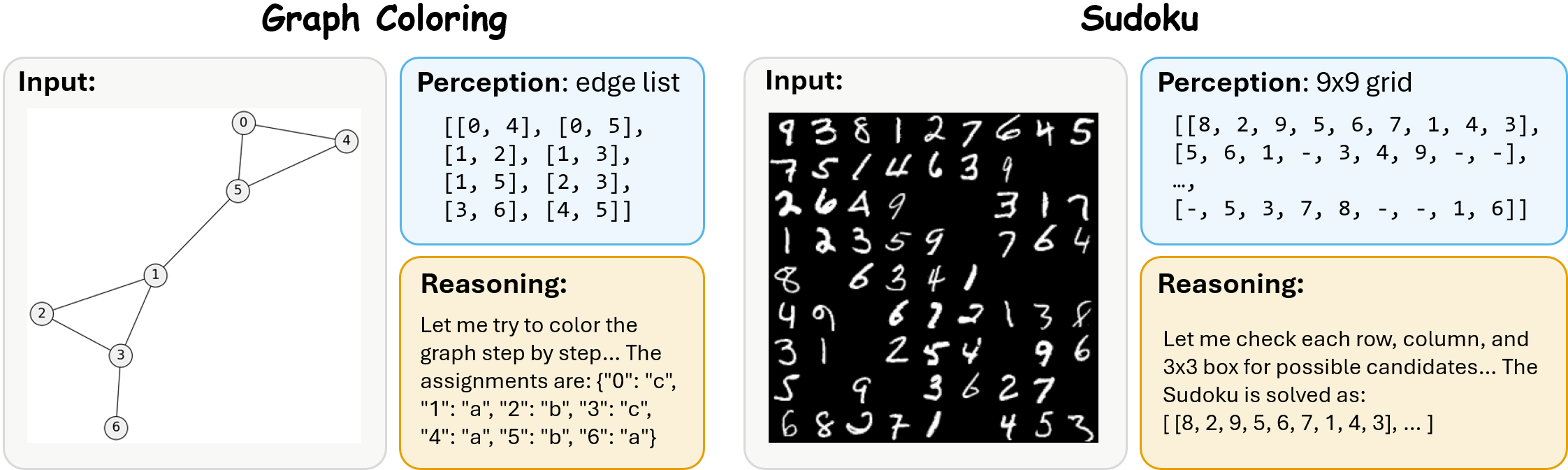}
    \vspace{-16pt}
    \caption{\small \textbf{Task illustration.} We show the graph coloring (\texttt{GC}) and \texttt{Sudoku} tasks with their input images, ground truth perception, and expected reasoning respectively.}
    \vspace{-12pt}
    \label{fig:tasks}
\end{figure*}

\subsection{Problem Setup}
\label{sec:problem_setup}

Let $\pi_\theta$ denote a vision-language model parameterized by $\theta$.
Given an input sequence $\mathbf{x}$ consisting of both visual and textual tokens, the model generates the output sequence $\mathbf{y} = [y_1,\cdots,y_T]$ in an auto-regressive manner, $y_t \sim \pi_\theta(\cdot\!\mid\!\mathbf{x}, \mathbf{y}_{<t})$.

\mypara{Disentangled outputs.}
Standard VLM CoT entangles perception with reasoning, making it difficult to supervise them separately or attribute errors. In practice, however, CoT traces often exhibit a natural order: models first verbalize the visual information and then reason over it. Recent work has started to explicitly enforce perception-before-reasoning format; this by itself improves final performance, and further enables targeted supervision or rewards on the perception components \citep{li2025selfrewardingvisionlanguagemodelreasoning,xia2025visionaryr1mitigatingshortcutsvisual,xiao2025perceptionr1advancingmultimodalreasoning,chen2025perception,wu2026seeing}.
Following this line of work, we structure each output $\mathbf{y}$ as two textual sequences, perception $\mathbf{p}$ followed by reasoning $\mathbf{r}$, separated by special tokens. Slightly abusing notation, we write the generation process as $\mathbf{p} \sim \pi_\theta(\cdot \!\mid\! \mathbf{x}), ~\mathbf{r} \sim \pi_\theta(\cdot \!\mid\! \mathbf{x}, \mathbf{p})$.

\mypara{Disentangled evaluation.} Our evaluation measures three metrics: end-to-end accuracy $a$, perception accuracy $a_p$, and reasoning accuracy $a_r$.

To support this evaluation, we assume the existence of a \textit{canonical textual representation} $\mathbf{p}^*$ that preserves the complete visual information needed for the task. This allows us to directly evaluate \textbf{perception accuracy} by comparing the model-generated perception $\mathbf{p}$ against the oracle perception:
$a_p=\mathbf{1}(\mathbf{p}=\mathbf{p}^*)$.
We further define a task-specific evaluator $Acc(\mathbf{r}\!\mid\!\cdot)$ that checks a reasoning output $\mathbf{r}$ over a textual perception representation rather than the raw image $\mathbf{x}$. The \textbf{end-to-end accuracy} is thus given by $a=Acc(\mathbf{r}\!\mid\! \mathbf{p}^*)$.

Evaluating reasoning accuracy is nontrivial because reasoning is conditioned on perception, so reasoning errors are naturally entangled with perception errors. An intuitive metric is \textbf{conditional reasoning accuracy}, $\Tilde{a}_r = Acc(\mathbf{r}\!\mid\!\mathbf{p})$, which evaluates whether the reasoning is correct under the model's own predicted perception. However, since $\mathbf{p}$ is model-generated, this metric does not isolate intrinsic reasoning ability: models may score higher by producing perceptions that are easier to reason from, rather than by reasoning more accurately. To decouple reasoning from perception, we instead use \textbf{counterfactual reasoning accuracy}. Specifically, we perform a \textit{counterfactual inference} by conditioning the model on the oracle perception and sampling $\mathbf{r}' \sim \pi_\theta(\cdot\!\mid\! \mathbf{x}, \mathbf{p}^*)$, and then measure $a_r=\mathrm{Acc}(\mathbf{r}'\!\mid\! \mathbf{p}^*)$. Unless otherwise stated, all reported reasoning accuracy refers to this counterfactual metric.

\subsection{Tasks Setup}

Our framework requires tasks with (i) a canonical perception representation $p^*$ for measuring perception accuracy, and (ii) a programmatic evaluator $Acc(\cdot \!\mid\! \cdot)$ defined solely over that representation. For real-world visual reasoning benchmarks, however, canonical perception is hard to define, and perception annotations are typically absent.
We therefore use \textbf{synthetic tasks as a controlled testbed}. Starting from text-only reasoning problems, we programmatically render visual inputs while retaining the original text input as the canonical perception representation $p^*$. As shown in Figure~\ref{fig:tasks}, we instantiate two tasks:
(1) \textbf{Graph Coloring} (GC), where $p^*$ is the graph edge list and correctness requires coloring all nodes with the graph's chromatic number while assigning different colors to adjacent nodes; and
(2) \textbf{Sudoku}, where $p^*$ is the partially filled $9 \times 9$ grid and correctness requires completing the grid such that every row, column, and $3 \times 3$ subgrid contains each digit exactly once.
As both tasks are programmatically constructed, we can synthesize large-scale training sets to perform controlled analysis.

\subsection{Training Paradigm}
\label{sec:training_formulation}

\mypara{Supervised fine-tuning.}
Given a dataset $\mathcal{D}=\{(\mathbf{x},\mathbf{y})\}$, where each input $\mathbf{x}$ is paired with a target output $\mathbf{y}$, SFT optimizes the token-averaged cross-entropy objective:
\vspace{-.5em}
\par
\vbox{\small
\begin{align*}
\mathcal{L}_{\mathrm{SFT}}
= - \mathbb{E}
\left[
\frac{1}{|\mathbf{y}|}
\sum_{t=1}^{|\mathbf{y}|}
\log \pi_\theta(y_t \!\mid\! \mathbf{x},\mathbf{y}_{<t})
\right].
\end{align*}}
\vspace{-1em}

\mypara{Reinforcement learning.} For RL, each input $\mathbf{x}$ is paired with a task reward
$R(\mathbf{y}\!\mid\! \mathbf{x})$, instantiated as the end-to-end accuracy in our setting. Following GRPO, for each input $\mathbf{x}$, we sample a group of $G$ responses
$\{\mathbf{y}^{(i)}\}_{i=1}^G$ from the behavior policy $\pi_{\theta_{\mathrm{old}}}$ and compute rewards
$r_i = R(\mathbf{y}^{(i)}\!\mid\! \mathbf{x})$. Rewards are normalized within the group to obtain a sequence-level advantage:
$\hat{A}_i =
\frac{r_i - \mathrm{mean}\left(\{r_j\}_{j=1}^G\right)}
{\mathrm{std}\left(\{r_j\}_{j=1}^G\right)}$, which is shared across all tokens in the response, i.e., $\hat{A}_{i,t}=\hat{A}_i$. Omitting clipping and KL regularization for clarity, the GRPO objective is:
\vspace{-.5em}
\par
\vbox{\small
\begin{align*}
&\mathcal{J}_\text{GRPO} = \mathbb{E} \left[ \frac{1}{G}
\sum_{i=1}^G
\frac{1}{|\mathbf{y}^{(i)}|}
\sum_{t=1}^{|\mathbf{y}^{(i)}|}
\rho_{i,t}(\theta)\hat{A}_{i,t} \right], 
\end{align*}}
\vspace{-.8em}
\noindent where $\rho_{i,t}(\theta)$
is the token-level importance ratio.

\subsection{Candidate Mechanisms and Interventions}
\label{sec:hypothesis_intervention}

The disentangled framework allows us to diagnose not only the presence of perception-reasoning asymmetry, but also its potential causes. We consider two candidate mechanisms: token imbalance in token-level objectives, and reward coupling when end-to-end accuracy is used as the RL reward. For each candidate mechanism, we formulate a testable hypothesis and introduce a targeted intervention, deferring empirical validation to \S\ref{sec:experiments}.

\mypara{H1: Token imbalance.} A first candidate mechanism is that perception receives a weaker optimization signal because it occupies a much smaller fraction of CoT tokens. In VLM CoT, perception is often a compact transcription of the visual input, while reasoning may span many tokens of verification, reflection, and self-correction. Prior work reports that perception accounts for only $17.6\%$ of tokens in base VLM outputs and drops to $12$--$14\%$ after RL training \citep{DBLP:journals/corr/abs-2510-08964}. As shown in \S\ref{sec:experiments:sft} and \S\ref{sec:experiments:rl}, a similar imbalance is present in our setting.

Since both SFT and RL use token-averaged objectives, perception and reasoning are implicitly weighted by their token counts. Taking SFT as an example,  the standard token-averaged loss can be decomposed as:
\vspace{-.5em}
\par
\vbox{\small
\begin{align*}
&\mathcal{L}_\text{SFT} = \frac{1}{|\mathbf{p}| + |\mathbf{r}|} \left( \mathcal{L}_p + \mathcal{L}_r \right),\\
&\mathcal{L}_p = -\mathbb{E} \left[\sum_{y_t\in\mathbf{p}} \log \pi_\theta (y_t\mid\mathbf{x},\mathbf{y}_{<t})\right], \\ 
&\mathcal{L}_r = - \mathbb{E}\left[\sum_{y_t\in\mathbf{r}} \log \pi_\theta (y_t\mid\mathbf{x},\mathbf{y}_{<t})\right],
\end{align*}}
\vspace{-.8em}
\noindent Thus, even when perception is necessary for solving the task, its shorter span gives it less aggregate influence on the update.

We test this hypothesis with \textbf{loss reweighting}. For SFT, we optimize against a reweighted loss,
$\mathcal{L}_{\text{SFT},\lambda} = \lambda \frac{\mathcal{L}_p}{|\mathbf{p}|}  + (1-\lambda) \frac{\mathcal{L}_r}{|\mathbf{r}|}$,
where $\lambda$ directly controls the relative emphasis on perception. When $\lambda$ exceeds the perception token fraction in the output CoT, perception is upweighted relative to standard training.
The same intervention can be applied to GRPO analogously, detailed in Appendix \ref{sec:appendix:token_reweighting_for_rl}. If token imbalance is a main driver of the asymmetry, increasing the perception weight should improve perception accuracy and reduce the end-to-end bottleneck.

\mypara{H2: Reward coupling.}
For RL, a second candidate mechanism is that the outcome reward, i.e., end-to-end accuracy $a$, is more weakly coupled to perception accuracy than to reasoning accuracy. Perception influences $a$ only when the downstream reasoning is also correct, whereas reasoning more directly determines the final answer and thus the reward. Consequently, using $a$ as the reward signal can induce noisy or misaligned credit assignment for perception, analogous to credit-assignment challenges for intermediate reasoning steps in text-only LLM RL \citep{lightman2024let}.

We test this hypothesis with \textbf{reward augmentation}. Instead of the standard outcome reward $R(y\!\mid\! x)=a$, we augment it with the perception accuracy $a_p$ and thus optimize %
$\mathcal{R}_\alpha(y\mid x)=\alpha \cdot a_p+(1-\alpha)\cdot a$,
where $\alpha\in[0,1]$ controls the strength of the perception signal. %
If reward coupling drives the RL asymmetry, increasing $\alpha$ should strengthen perception optimization and mitigate a perception-reasoning imbalance.

\section{Experimental Results}
\label{sec:experiments}

We empirically investigate four questions: (1) Does post-training optimize perception and reasoning asymmetrically? (2) What mechanism drives the asymmetry? (3) Which interventions best re-balance perception and reasoning, and how do they affect end-to-end accuracy? (4) Do the identified interventions transfer to real-world visual reasoning tasks? In this section, we present results addressing questions (1)--(3) for SFT and RL in \S\ref{sec:experiments:sft} and \S\ref{sec:experiments:rl}, respectively, and  evaluate transfer to real-world tasks in \S\ref{sec:experiments:realworld}.

\subsection{Settings for Controlled Experiments}
\label{sec:experiments:settings}

We experiment with \texttt{Qwen3-VL-2B-Instruct} \citep{bai2025qwen3vltechnicalreport} and \texttt{InternVL3.5-2B} \citep{wang2025internvl3} on the \texttt{GC} and \texttt{Sudoku} tasks. For each task, we generate $500/2k/16k$ data instances for testing, SFT training, and RL training respectively. To curate  supervision for SFT, we concatenate ground-truth perception with reasoning distilled from \texttt{Qwen3-32B}. For loss reweighting, we experiment with $\lambda \in \{0.1,0.2,0.5,0.8,0.9\}$; for reward coupling, we experiment with $\alpha \in \{0.0,0.2,0.5,0.8,1.0\}$, where $\alpha\!=\!0$ corresponds to the standard RL. Full details are in Appendix~\ref{sec:appendix:exp_settings}.

\begin{figure*}[!tbph]
    \centering
    \begin{minipage}[b]{0.43\textwidth}
    \centering
        \includegraphics[width=\linewidth]{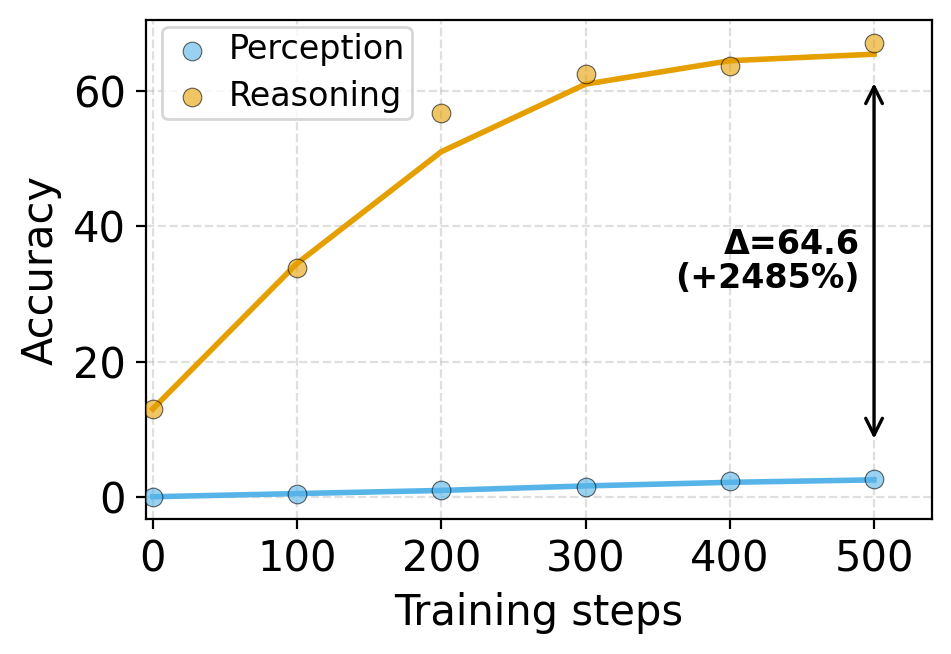}
    \vspace{-2.2em}
        \caption{\small Perception and reasoning accuracy during SFT optimization of \texttt{Qwen} model on \texttt{GC} task, highlighting the asymmetric optimization.}
        \label{fig:sft_gap}
    \end{minipage}
    \hfill
    \begin{minipage}[b]{0.54\textwidth}
        \centering
        \includegraphics[width=\linewidth]{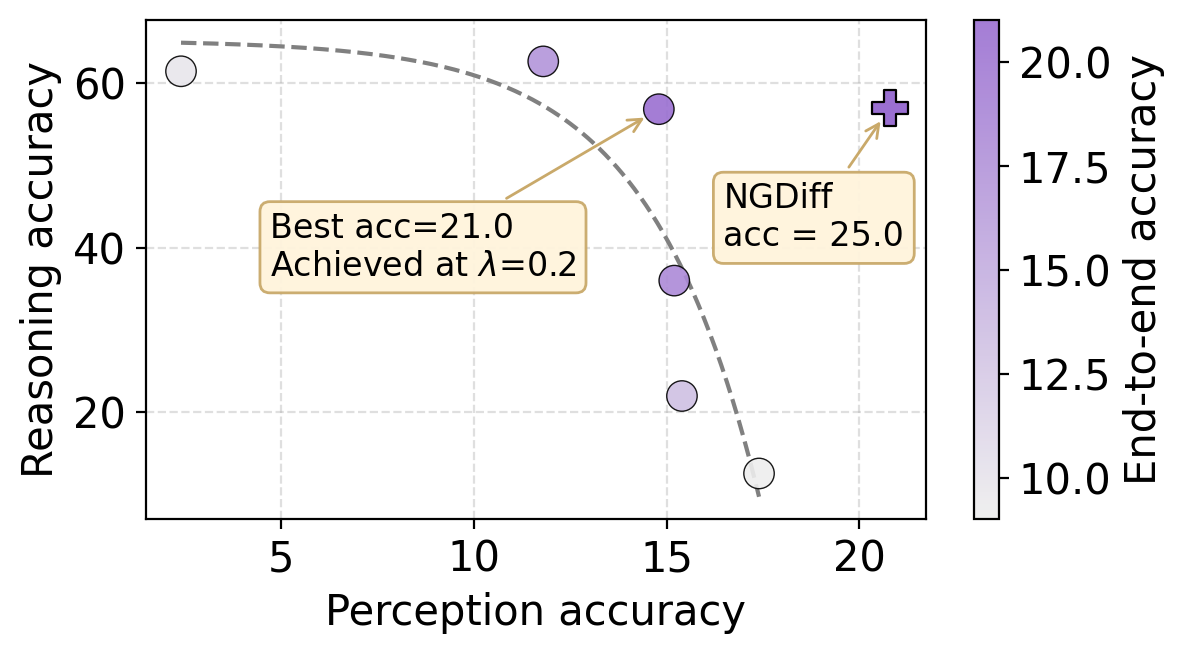}
    \vspace{-2.2em}
        \caption{\small Perception-reasoning trade-off when training \texttt{Qwen} on the \texttt{GC} task with SFT by varying $\lambda$. We also show that NGDiff achieves a better trade-off and higher end-to-end accuracy.}
        \label{fig:sft_tradeoff}
    \end{minipage}
    \vspace{-.9em}
\end{figure*}

\begin{figure}[!tbhp]
    \centering
    \vspace{-.2em}
    \includegraphics[width=\linewidth]{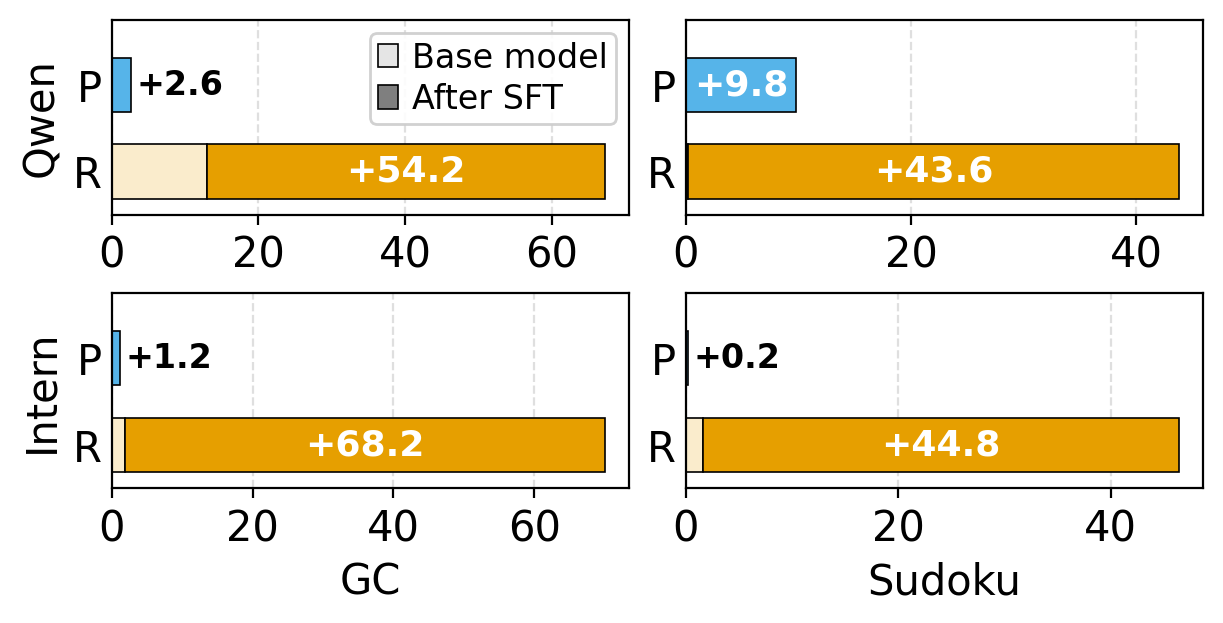}
    \vspace{-2.2em}
    \caption{\small \textbf{Perception and reasoning accuracy before and after SFT.} Both start near zero, but SFT substantially improves reasoning while leaving perception largely flat.}
    \vspace{-1.1em}
    \label{fig:sft_gap_bar}
\end{figure}

\subsection{Supervised Fine-Tuning Phase}
\label{sec:experiments:sft}

\mypara{Asymmetric optimization in SFT.}
We first investigate how standard SFT optimizes perception and reasoning performance.
Starting from base models that achieve near-zero accuracy in both perception and reasoning on our out-of-distribution tasks, SFT exhibits a pronounced asymmetry: it substantially improves reasoning while yielding only limited gains in perception.
This pattern is consistent across all four model-task settings, as summarized in Figure~\ref{fig:sft_gap_bar}; Figure~\ref{fig:sft_gap} further illustrates the optimization dynamics of \texttt{Qwen} on the \texttt{GC} task as a representative setting.

\mypara{Testing H1: Token imbalance drives SFT asymmetry.}
We first test whether SFT asymmetry can be explained by token imbalance in CoT supervision. Since the reasoning traces are distilled from a text-only RL model, following common practice \citep{deng2025openvlthinker}, they occupy substantially more tokens than the perception component: perception accounts for only $2.2\%$ of tokens in \texttt{GC} and $2.5\%$ in \texttt{Sudoku}. This imbalance yields a much weaker optimization signal: in the \texttt{Qwen}-\texttt{GC} setting, for example, perception contributes only $1.3\%$ of the loss and $8.5\%$ of the gradient norm, supporting token imbalance as a concrete mechanism behind SFT asymmetry.

We then employ the loss reweighting intervention in \S\ref{sec:hypothesis_intervention} as a mechanism test. As shown in Figure~\ref{fig:sft_tradeoff}, varying $\lambda$ reveals a Pareto-like frontier: under a fixed training budget, increasing $\lambda$ improves perception accuracy but weakens reasoning optimization. Importantly, \textbf{the best end-to-end visual reasoning accuracy is achieved by moderately upweighting perception}. As reported in Table~\ref{tab:sft_main}, the best $\lambda$ improves end-to-end performance over standard SFT by $10.0$--$13.8$ points across all four settings, suggesting loss reweighting as a simple and effective mitigation for SFT asymmetry.

\begin{table}[!t]
\renewcommand{\arraystretch}{1.1}
\small
    \centering
    \begin{tabular}{l|rr|rr}
    \hline
& \multicolumn{2}{c|}{GC} & \multicolumn{2}{c}{Sudoku} \\[-2pt]
& \multicolumn{1}{c}{\scriptsize \texttt{Qwen}} & \multicolumn{1}{c|}{\scriptsize \texttt{InternVL}} & \multicolumn{1}{c}{\scriptsize \texttt{Qwen}} & \multicolumn{1}{c}{\scriptsize \texttt{InternVL}} \\
\hline
Standard SFT & 9.8 & 10.6 & 6.4 & 0.8 \\
\hline
Loss Reweighting & 21.0 & \textbf{20.6} & 18.4 & 14.6 \\[-2pt]
& {\scriptsize \color{mygreen}$\uparrow$11.2} & {\scriptsize \color{mygreen}$\uparrow$10.0} & {\scriptsize \color{mygreen}$\uparrow$12.0} & {\scriptsize \color{mygreen}$\uparrow$13.8} \\
NGDiff & \textbf{25.0} & \textbf{20.6} & \textbf{22.0} & \textbf{19.0} \\[-2pt]
& {\scriptsize \color{mygreen}$\uparrow$15.2} & {\scriptsize \color{mygreen}$\uparrow$10.0} & {\scriptsize \color{mygreen}$\uparrow$15.6} & {\scriptsize \color{mygreen}$\uparrow$18.2} \\
\hline
    \end{tabular}
    \vspace{-.6em}
    \caption{\small \textbf{End-to-end accuracy after SFT.} We compare standard SFT and two SFT asymmetry mitigation methods: loss reweighting with the best tuned $\lambda$ and NGDiff.}
    \vspace{-1.2em}
    \label{tab:sft_main}
\end{table}

\mypara{Dynamic weighting further improves the trade-off.} While loss reweighting improves end-to-end performance, its optimal $\lambda$ is model- and task-dependent, requiring expensive hyperparameter tuning. We thus evaluate NGDiff that dynamically balances multiple objectives based on their gradient norms \citep{jin2025unlearning}. For each mini-batch, let $\mathbf{g}_p$ and $\mathbf{g}_r$ denote the gradients from the perception and reasoning objectives, respectively. NGDiff then sets the perception weight as $\lambda = \frac{1}{||\mathbf{g}_p||} ~/ ~\left(\frac{1}{||\mathbf{g}_p||} + \frac{1}{||\mathbf{g}_r||} \right)$. As shown in Table~\ref{tab:sft_main}, NGDiff consistently matches or outperforms the manually tuned loss-reweighting in all settings, with gains of up to $4.4$ points. This highlights NGDiff as a practical and effective mitigation for SFT asymmetry.

\vspace{.1em} %

\begin{takeawaybox}{Takeaway for SFT}
\textbf{Findings:} SFT under-optimizes perception because reasoning tokens dominate the CoT trace, leaving perception weakly optimized and bottlenecking end-to-end performance. \\[.3em] %
\textbf{Suggestions: Reweight the SFT loss toward perception.}  Moderate perception upweighting improves end-to-end performance, while NGDiff provides a tuning-free way to approach a better perception-reasoning balance.
\end{takeawaybox}

\vspace{.1em} %

\subsection{Reinforcement Learning Phase}
\label{sec:experiments:rl}

\begin{figure}[!t]
    \centering
    \includegraphics[width=\linewidth]{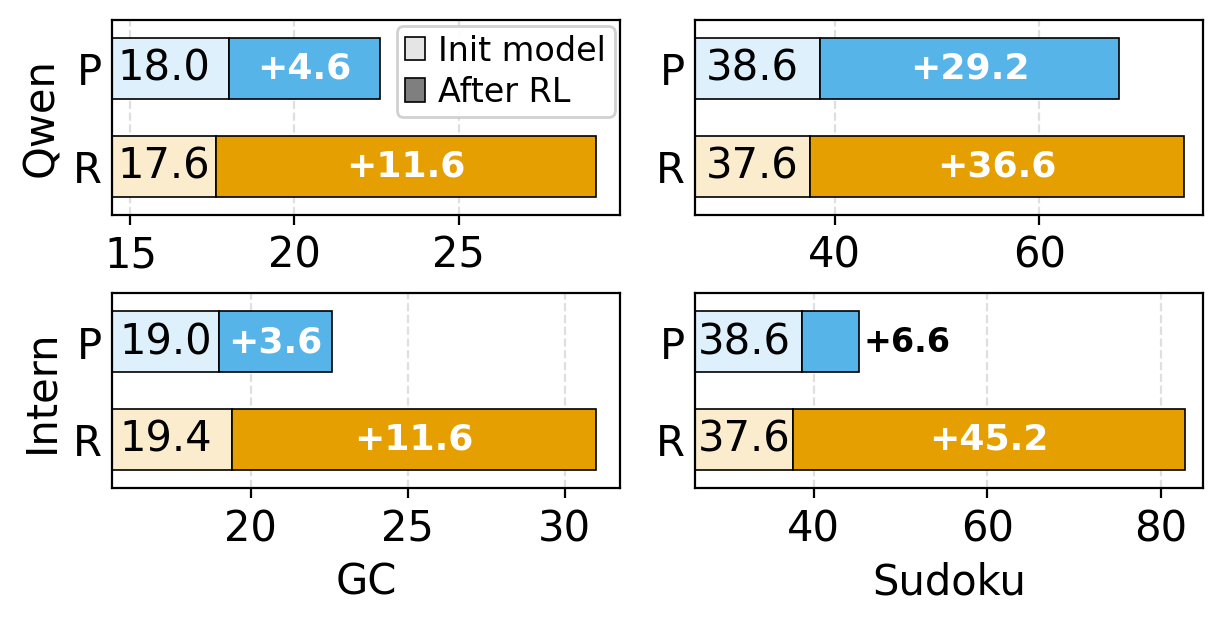}
    \vspace{-2.2em}
    \caption{\small \textbf{Perception and reasoning accuracy before and after RL}, where RL improves reasoning more substantially than perception.}
    \vspace{-1em}
    \label{fig:rl_gap_bar}
\end{figure}

\begin{figure*}[!tbph]
    \centering
    \begin{minipage}[b]{0.43\textwidth}
        \centering
        \includegraphics[width=\textwidth]{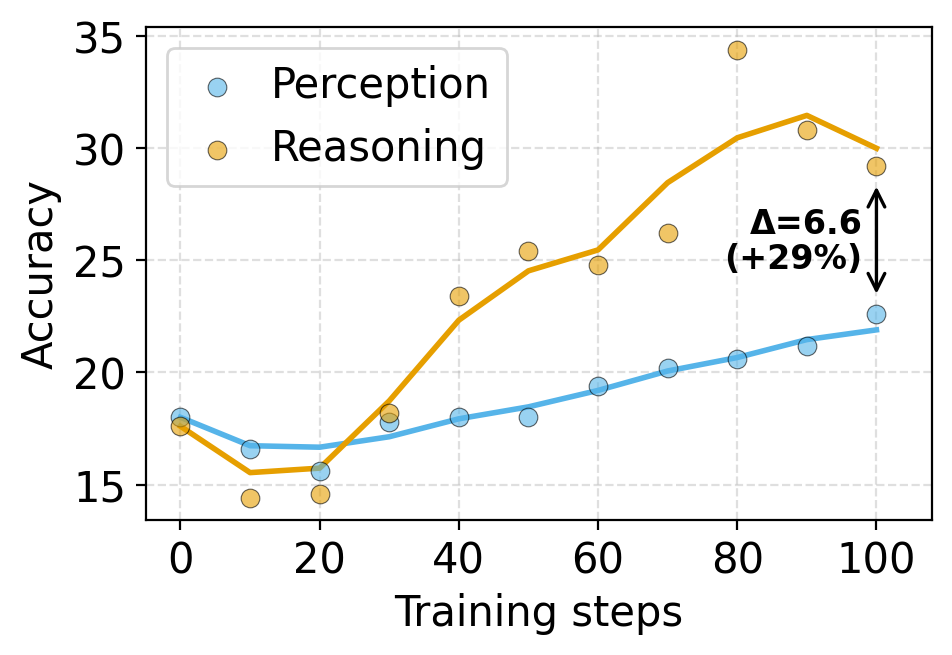}
    \vspace{-2.35em}
        \caption{\small Perception and reasoning accuracy during RL optimization of \texttt{Qwen} model on \texttt{GC} task.}
        \label{fig:rl_gap}
    \end{minipage}
    \hfill
    \begin{minipage}[b]{0.54\textwidth}
        \centering
        \includegraphics[width=\linewidth]{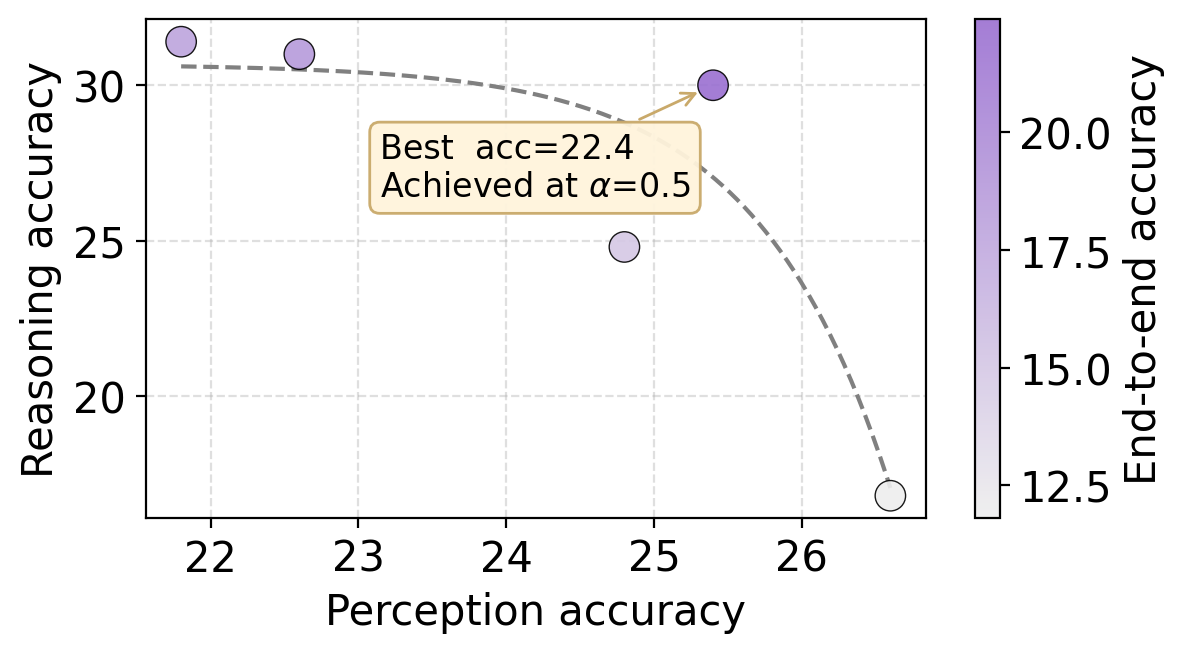}
    \vspace{-2.35em}
        \caption{\small Perception-reasoning trade-off when training \texttt{Qwen} on the \texttt{GC} task with RL by varying $\alpha$.}
        \label{fig:rl_tradeoff}
    \end{minipage}
    \vspace{-.4em}
\end{figure*}

\mypara{Asymmetric optimization in RL.} We next investigate how RL
optimizes perception and reasoning. To ensure a fair comparison, we initialize RL from SFT checkpoints with matched perception and reasoning accuracies. As shown in Figure~\ref{fig:rl_gap_bar}, RL exhibits a similar asymmetry to SFT: it consistently improves reasoning more than perception across all four settings. Figure~\ref{fig:rl_gap} further shows the optimization dynamics for the \texttt{Qwen}-\texttt{GC} setting as a representative case.

\mypara{Testing H1: token imbalance does not explain RL asymmetry.}
Since RL is initialized from SFT checkpoints, the CoT token imbalance persists and even intensifies: in \texttt{Qwen}-\texttt{GC}, for example, the perception-token fraction drops from $2.2\%$ to $1.5\%$ after RL training. However, the reweighting intervention from \S\ref{sec:hypothesis_intervention} no longer mitigates the asymmetry. Standard RL achieves $a/a_p/a_r = 18.8/22.6/31.0$, while reweighting policy-gradient terms with $\lambda=0.1$, $0.2$, and $0.5$ degrades performance to $17.8/15.4/29.8$, $17.4/14.2/26.6$, and $13.2/2.2/18.8$, respectively. Thus, although token imbalance remains, directly upweighting perception in the RL objective consistently disrupts training rather than improving perception, indicating that RL asymmetry is not primarily driven by token allocation.

\begin{figure}[!tb]
    \centering
    \includegraphics[width=\linewidth]{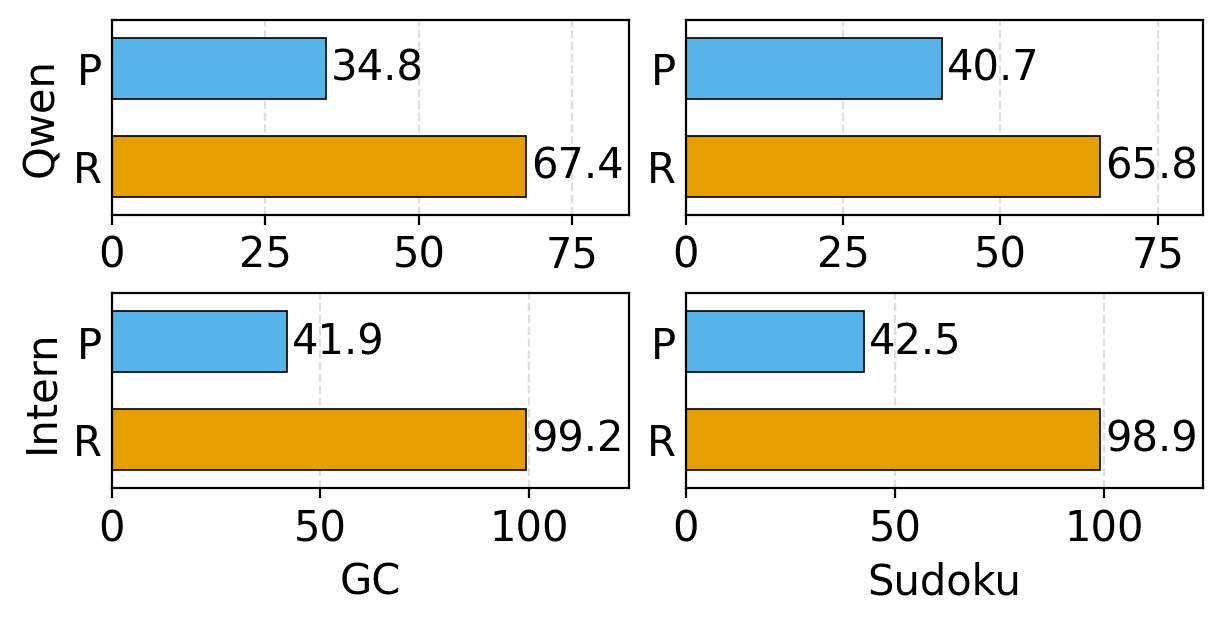}
    \vspace{-2.2em}
    \caption{\small Correlation between outcome reward and perception (P) or reasoning (R) accuracy, measured by Pearson's $r$.}
    \vspace{-.5em}
    \label{fig:rl_correlation_single}
\end{figure}

\mypara{Testing H2: Reward coupling drives RL asymmetry.}
To quantitatively test reward coupling as a potential cause, we measure how strongly the outcome reward aligns with perception $vs.$ reasoning correctness. For each setting, we sample $2500$ rollouts, evaluate the perception and reasoning correctness for each rollout, and compute Pearson's $r$ between each correctness metric and the outcome reward. For perception, we use perception accuracy $a_p$. For reasoning, we use conditional reasoning accuracy $\Tilde{a}_r = Acc(\mathbf{r}\mid\mathbf{p})$, since counterfactual reasoning accuracy requires resampling reasoning under oracle perception and is therefore not defined within a single rollout. Figure~\ref{fig:rl_correlation_single} shows that the outcome reward is consistently more correlated with reasoning than with perception, indicating that standard RL supplies a stronger learning signal for reasoning and supporting reward coupling as the primary driver of RL asymmetry.

\begin{figure*}
        \centering
        \includegraphics[width=\linewidth]{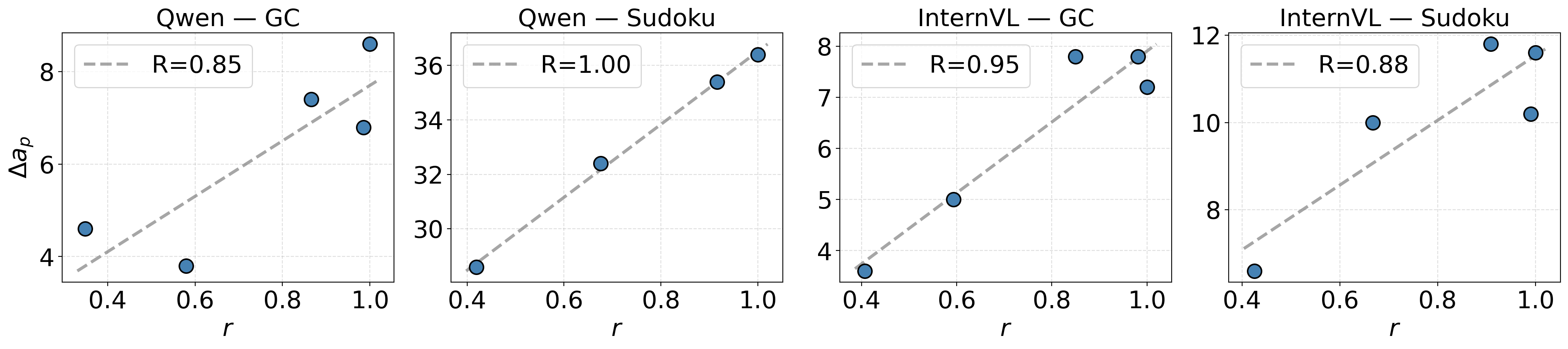}
    \vspace{-2.3em}
        \caption{\small The correlation $r$ between RL reward and perception accuracy reliably predicts the perception gains $\Delta a_p$ after RL.}
    \vspace{-.95em}
        \label{fig:correlation_performance}
\end{figure*}

\begin{table}[!t]
\renewcommand{\arraystretch}{1.1}
\setlength{\tabcolsep}{4pt}
\small
    \centering
    \begin{tabular}{l|rr|rr}
    \hline
& \multicolumn{2}{c|}{GC} & \multicolumn{2}{c}{Sudoku} \\[-2pt]
& \multicolumn{1}{c}{\scriptsize \texttt{Qwen}} & \multicolumn{1}{c|}{\scriptsize \texttt{InternVL}} & \multicolumn{1}{c}{\scriptsize \texttt{Qwen}} & \multicolumn{1}{c}{\scriptsize \texttt{InternVL}} \\
\hline
Init model & 10.2 & 15.8 & 18.4 & 14.6 \\
\hline
Standard RL $(\alpha\!=\!0)$ & 18.8 & 20.6 & 50.6 & \textbf{40.2} \\
Reward Augmentation & \textbf{22.4} & \textbf{22.6} & \textbf{56.6} & 38.6 \\[-2pt]
        & {\scriptsize \color{mygreen}$\uparrow$3.6}
        & {\scriptsize \color{mygreen}$\uparrow$2.0}
        & {\scriptsize \color{mygreen}$\uparrow$6.0}
        & {\scriptsize \color{red}$\downarrow$1.6} \\
\hline
    \end{tabular}
    \vspace{-.8em}
    \caption{\small \textbf{End-to-end accuracy after RL}. We compare standard RL against the reward augmentation mitigation approach with the best tuned non-zero $\alpha$.}
    \vspace{-1.7em}
    \label{tab:rl_main}
\end{table}

We then employ reward augmentation as a mechanism test. Varying $\alpha$ exposes a perception-reasoning trade-off as shown in Figure \ref{fig:rl_tradeoff}. Although the frontier is noisier than in SFT due to RL variance, the trend is consistent: larger $\alpha$ improves perception but weakens reasoning under the same training budget. Figure \ref{fig:correlation_performance} shows that across different $\alpha$ settings, reward-perception correlation reliably predicts the perception gains after RL, suggesting \textbf{reward-perception correlation as a useful diagnostic} for reward design. Finally, Table \ref{tab:rl_main} further shows that end-to-end accuracy is  maximized at an intermediate, non-zero $\alpha$ in most settings, indicating \textbf{perception-aware reward augmentation is an effective mitigation}.

\mypara{Surrogate perception rewards offer a practical approximation.}
Since ground-truth perception rewards are rarely available in realistic settings, we evaluate three surrogate designs from prior work: CLIP-based image-text similarity~\citep{chen2025perception}, self-rewarding with rollout budgets $N\in\{1,4\}$~\citep{li2025selfrewardingvisionlanguagemodelreasoning}, and teacher-based rewards using \texttt{gemini-3-flash-preview}~\citep{xiao2025perceptionr1advancingmultimodalreasoning}. Implementation details are in Appendix \ref{sec:appendix:surrogate_rewards_impl}. Using correlation with ground-truth perception accuracy as a diagnostic, Table~\ref{tab:surrogate_reward} shows that CLIP provides little useful signal, whereas the teacher reward is consistently the most reliable. Self-rewarding with $N=1$ only modestly improves over the outcome reward, while increasing the budget to $N=4$ substantially strengthens the signal at additional computational cost.

\begin{table}[!t]
\renewcommand{\arraystretch}{1.1}
    \centering
    \small
    \begin{tabular}{l|rr|rr}
\hline
& \multicolumn{2}{c|}{GC} & \multicolumn{2}{c}{Sudoku} \\[-2pt]
& \multicolumn{1}{c}{\scriptsize \texttt{Qwen}} & \multicolumn{1}{c|}{\scriptsize \texttt{InternVL}} & \multicolumn{1}{c}{\scriptsize \texttt{Qwen}} & \multicolumn{1}{c}{\scriptsize \texttt{InternVL}} \\
\hline
Outcome Reward & 34.8 & 40.7 & 41.9 & 42.5 \\
\hline
CLIP Similarity & -0.1 & -0.1 & 5.8  &  8.0  \\
Self Reward (N=1) & 32.0 & 37.3 & 46.6 & 45.7 \\
Self Reward (N=4) & 65.8 & 67.1 & 71.3 & 69.8 \\
Teacher Reward & \textbf{96.0} & \textbf{95.6} & \textbf{74.9} & \textbf{72.0} \\
\hline
\end{tabular}
    \vspace{-.8em}
\caption{\small Pearson's correlation $r$ between perception accuracy and different surrogate rewards.} %
    \vspace{-.5em}
    \label{tab:surrogate_reward}
\end{table}

\begin{table}[!t]
\renewcommand{\arraystretch}{1.1}
    \centering
    \small
    \begin{tabular}{l|r|rrr}
\hline
& $r$ & $a$ & $a_p$ & $a_r$ \\
\hline
Outcome only & 34.8 & 18.8 & 22.6 & 31.0 \\
\hline
\rowcolor{gray!5}
\multicolumn{5}{l}{%
  \rule{0pt}{2.2ex}\textit{Reward Augmentation ($\alpha\!=\!0.5$)}%
}\\[.5pt]
Self Reward (N=1) & 32.0 & 18.4 & 21.8 & 27.6 \\
Teacher reward & \underline{96.0} & \underline{21.0} & \underline{25.0} & \textbf{30.6} \\
GT perception reward & \textbf{100.0} & \textbf{22.4} & \textbf{25.4} & \underline{30.0} \\
\hline
\end{tabular}
    \vspace{-.5em}
\caption{\small \textbf{RL performance under different reward settings} in the \texttt{Qwen}-\texttt{GC} setting. For each reward, we report its Pearson correlation $r$ with perception accuracy. $a/a_p/a_r$ denote end-to-end, perception, and reasoning accuracies, respectively.}
    \vspace{-1.4em}
    \label{tab:rl_surrogate}
\end{table}

We further validate this diagnostic with a proof-of-concept RL run on \texttt{Qwen}-\texttt{GC}. With $\alpha\!=\!0.5$, we augment the outcome reward using self-rewarding ($N=1$) or teacher-based perception rewards, and compare both against standard RL as well as the oracle perception reward. As shown in Table~\ref{tab:rl_surrogate}, stronger reward-perception correlation yields larger perception gains: the teacher reward nearly matches the oracle reward and clearly outperforms self-rewarding. These results support \textbf{reward-perception correlation as a practical diagnostic} for surrogate design, and suggest  \textbf{surrogate perception rewards as practical mitigations} when sufficiently aligned with ground-truth perception.

\vspace{.1em} %

\begin{takeawaybox}{Takeaway for RL}
\textbf{Findings:} RL under-optimizes perception because outcome rewards are more tightly coupled with reasoning than with perception, yielding a weak signal for perception optimization. \\[.3em] %
\textbf{Suggestions: Add perception-aware rewards.} Moderately augmenting the outcome reward with a perception reward improves end-to-end performance. When oracle perception rewards are unavailable, reliable surrogate rewards also yield gains, with reward-perception correlation as a simple diagnostic for reward reliability.
\end{takeawaybox}

\vspace{.1em} %

\subsection{Transfer to Real-World  Tasks}
\label{sec:experiments:realworld}

Having identified targeted mitigations in \S\ref{sec:experiments:sft} and \S\ref{sec:experiments:rl}, we next investigate whether their benefits generalize beyond the controlled setting to real-world visual reasoning tasks. Specifically, we train \texttt{Qwen3-VL-2B-Instruct} on the SFT and RL data from Vision-SR1 \citep{li2025selfrewardingvisionlanguagemodelreasoning}, which provides separate annotations for perception and reasoning. We use the same training hyperparameters as in our prior experiments, reported in Tables \ref{tab:appendix:sft_hyperpara} and \ref{tab:appendix:qwenvl_rl_hyperpara}. We evaluate the resulting model on three real-world benchmarks: HallusionBench \citep{guan2024hallusionbench}, MMMU-Pro \citep{yue2025mmmu}, and two vision-heavy subsets of MathVerse  (Vision-Dominant and Vision-Only) \citep{zhang2024mathverse}.

\mypara{SFT results.} Compared to standard SFT, NGDiff yields a \textbf{+3.3} improvement on HallusionBench (64.5 vs.\ 61.2) and a \textbf{+1.3} gain on MMMU-Pro (23.9 vs.\ 22.6), while maintaining the same performance on MathVerse (27.4), all without requiring tuning of the hyperparameter $\lambda$. These results demonstrate that the benefits of NGDiff transfer beyond the controlled setting.

\mypara{RL results.} Starting from the standard SFT model, we further train with both standard RL and perception-aware reward augmentation using a \textit{teacher-based} surrogate reward. Specifically, we prompt \texttt{Qwen3-VL-30B-A3B-Instruct} to judge whether the perception is faithful to the image and sufficient for answering the question, with details in Appendix \ref{sec:appendix:surrogate_rewards_impl}. We fix $\alpha\!=\!0.5$ without further tuning, as it proved effective in the synthetic settings. Perception-aware reward improves performance on MathVerse (28.7 vs.\ 27.4) and MMMU-Pro (25.1 vs.\ 24.2), while maintaining the same performance on HallusionBench (63.6). This supports the transferability of perception-aware RL and shows that a teacher-based surrogate reward can provide useful perception supervision in realistic settings.

\section{Related Work}
\label{sec:related_work}

\mypara{Post-training in VLMs.} Post-training is known to improve reasoning capabilities in pre-trained VLMs, traditionally via supervised fine-tuning on long CoT \citep{xu2025llava,yang2025r1,yao2026mulberry}. More recently, the success of GRPO in LLM reasoning \citep{shao2024deepseekmath,guo2025deepseek,jaech2024openai} has motivated its adaptation to VLMs \citep{huang2025vision,peng2025lmmr1,meng2025mm,zhang2025r1,bai2025qwen3vltechnicalreport}. However, stronger reasoning does not necessarily translate into more reliable perception. VLMs are known to suffer from hallucination \citep{li2023evaluating,bai2024hallucination,liu2024survey}; recent work finds that, despite improving reasoning, RL yields little improvement in perception and may even degrade it \citep{liu2025more,tian2025more,yao2025rethinking,xiao2025perceptionr1advancingmultimodalreasoning,wang2025perception}. Together, these findings point to perception as a key bottleneck in VLM post-training.

\mypara{Mitigation approaches for perception.} A growing body of work attempts to mitigate the perception bottleneck. On the SFT side, several work has explored augmenting CoT with perception 
\citep{shao2024visual, wang2025vgr,man2025argus,xu2025llava,DBLP:journals/corr/abs-2510-08964}, but their customized chain-of-thought designs limit their application to generic post-training. On the RL side, mitigations largely focus on perception-aware reward shaping \citep{xia2025visionaryr1mitigatingshortcutsvisual,xiao2025perceptionr1advancingmultimodalreasoning,li2025selfrewardingvisionlanguagemodelreasoning,wang2025perception,chen2025perception}. In \S\ref{sec:experiments:rl}, we study representative perception-reward designs under our diagnostic framework, and show their transfer to real-world tasks in \S\ref{sec:experiments:realworld}.

\mypara{Controlled analysis of post-training.} Our methodology follows a line of work using controlled synthetic tasks to remove confounders  in real-world benchmarks and dissect  model training. For example, \citet{chusft} use synthetic settings to compare SFT memorization with RL generalization; \citet{li2025unveiling} probe compositional generalization; and \citet{li2026doesrlimprovevisual} localize the internal changes induced by RL. We extend this line of work with two synthetic tasks that disentangle the optimization of reasoning and perception in post-training.

\section{Conclusions}

This work investigates asymmetric optimization in VLM post-training, where reasoning improves more reliably than perception, causing perception to become a bottleneck for end-to-end performance. Using a controlled experimental framework, we identify a consistent perception-reasoning asymmetry for both SFT and RL training and trace it to different underlying mechanisms. In SFT, the asymmetry is primarily driven by token imbalance, which can be mitigated through  loss reweighting. In RL, the asymmetry instead arises mainly from reward coupling, and incorporating ground-truth perception rewards or reliable surrogate rewards effectively reduces this imbalance. These interventions are not only effective in controlled synthetic settings but also transfer to real-world scenarios, yielding improvements across three benchmarks. Overall, our findings diagnose a pronounced asymmetry under standard post-training objectives and suggest perception-aware interventions for strengthening perception learning and improving end-to-end visual reasoning performance.

\section*{Limitations}

This work studies whether and why reasoning and perception improve differently during VLM post-training primarily through two synthetic visual reasoning tasks. A limitation of this design is that synthetic settings abstract away from the full diversity and complexity of real-world vision-language data. At the same time, this abstraction is central to our analysis: our goal is to isolate the training dynamics that cause perception and reasoning to be optimized asymmetrically, rather than to benchmark end-task performance. Synthetic tasks provide a particularly clean setting for this purpose, enabling full control over the input distribution, canonical perception annotations, and reliable programmatic evaluation. Without such control, perception and reasoning errors are often entangled, making it difficult to measure each component independently or identify the mechanisms responsible for their divergence.

As discussed in \S\ref{sec:related_work}, controlled synthetic settings are widely used to study model behavior while minimizing confounding factors. Moreover, the mechanisms we identify---token imbalance in SFT and reward coupling in RL---arise from general properties of post-training objectives rather than from the surface form of our synthetic tasks. To further assess the practical relevance of these findings, we evaluate the corresponding mitigation strategies on real-world visual reasoning benchmarks in \S\ref{sec:experiments:realworld}. While these benchmarks do not support the same disentangled diagnosis, the improvements from NGDiff and perception-aware reward augmentation provide complementary evidence that the interventions identified through controlled analysis can transfer beyond synthetic settings. Nevertheless, our mechanistic conclusions are established in controlled tasks, and extending the same fine-grained diagnosis to more diverse real-world data remains an important direction for future work.

\bibliography{custom}

\appendix

\section{Responsible Research Checklist}

\subsection{Used Scientific Artifacts}

\begin{enumerate}[itemsep=0pt, topsep=0pt, parsep=0pt, partopsep=0pt]
    \item \texttt{Qwen/Qwen3-VL-2B-Instruct} \citep{bai2025qwen3vltechnicalreport} for post-training, released under Apache-2.0 license;
    \item \texttt{OpenGVLab/InternVL3\_5-2B} \citep{wang2025internvl3} for post-training, released under Apache-2.0 license;
    \item \texttt{LlamaFactory} \citep{zheng2024llamafactory} for SFT training, released under Apache-2.0 license;
    \item \texttt{verl} \citep{sheng2024hybridflow} for training the \texttt{Qwen3-VL} model with RL, released under Apache-2.0 license;
    \item \texttt{verl-internvl} \citep{verl_internvl_github} for training the \texttt{InternVL3.5} model with RL, released under Apache-2.0 license;
    \item \texttt{vllm} \citep{kwon2023efficient} for inference and evaluation, released under Apache-2.0 license;
    \item The graph coloring dataset \citep{heyman2025evaluating} released under MIT license is adapted for building our graph coloring task;
    \item The Sudoku dataset \citep{du2024learning} released under MIT license is adapted for building our Sudoku task;
    \item \texttt{MNIST} images \citep{lecun2010mnist} released under MIT license for rendering Sudoku images;
    \item SFT and RL data from Vision-SR1 \citep{li2025selfrewardingvisionlanguagemodelreasoning}, released under Apache-2.0 license;
    \item HallusionBench dataset \citep{guan2024hallusionbench}, released under BSD-3-Clause license.
    \item MMMU-Pro dataset \citep{yue2025mmmu}, released under Apache-2.0 license;
    \item MathVerse dataset \citep{zhang2024mathverse}, released under MIT license.
\end{enumerate}
We have reviewed the licenses and intended-use conditions of all existing artifacts we used. Our use is fully within a research context: the models, frameworks, and datasets are used only for research-oriented post-training, supervised fine-tuning, reinforcement learning, and benchmark construction.

\subsection{Created Scientific Artifacts}

\mypara{Code.} The code artifacts include: (1) training code for our targeted interventions, built on top of \texttt{LlamaFactory} and \texttt{verl}; and (2) detailed scripts for data processing, supervised fine-tuning, reinforcement learning, evaluation, and result reproduction.

\mypara{Datasets.} The dataset artifacts cover both \texttt{GC} and \texttt{Sudoku}. Each task includes an SFT training dataset, an RL training dataset, and a test set. All examples are in English. Since the tasks are synthetic or puzzle-based, they do not represent natural-language demographic groups, human subjects, or personally identifiable information. Therefore, demographic coverage is not applicable. Statistics are reported in Table~\ref{tab:created_artifact_stats}.

\begin{table}[!h]
\centering
\small
\begin{tabular}{l|ccc}
\toprule
\textbf{Task} & \textbf{\# Test} & \textbf{\# SFT Train} & \textbf{\# RL Train} \\
\midrule
\texttt{GC} & 500 & 2,000 & 16,000 \\
\texttt{Sudoku} & 500 & 2,000 & 16,000 \\
\bottomrule
\end{tabular}
\caption{Statistics for the created dataset artifacts. The tasks are synthetic reasoning tasks in English and do not involve human subjects, demographic groups, or personally identifiable information.}
\label{tab:created_artifact_stats}
\end{table}

For \texttt{GC}, we curate random graphs with 7, 8, or 9 nodes and render them as images using \texttt{networkx}. For \texttt{Sudoku}, we adapt the problem generation code from \citep{du2024learning} and render images by combining MNIST digit images \citep{lecun2010mnist}. Detailed dataset synthesis scripts will also be released under the MIT license.

\mypara{Release plan.}
We will release all artifacts upon publication. The code will be released under the Apache-2.0 license, and the data will be released under the MIT license. For components adapted from existing artifacts, we will preserve the required copyright, license, and attribution notices.

\subsection{AI Assistants In Research Or Writing}

I used AI assistants solely for stylistic improvements in writing, such as improving clarity, grammar, and phrasing. I did not use AI assistants for coding, brainstorming, research design, data analysis, interpretation of results, or any other critical intellectual contribution.

\section{Experimental Details}
\label{sec:appendix:exp_settings}

\mypara{SFT training.}
We perform SFT using the \texttt{LlamaFactory} package. The hyperparameter settings are reported in Table~\ref{tab:appendix:sft_hyperpara}. Each experiment is performed on 4 $\times$ 80GB A100 GPUs with \texttt{per\_device\_train\_batch\_size} set to 2 and \texttt{gradient\_accumulation\_steps} set to 1, giving an effective batch size of 8. Each run takes approximately 3 hours.

\begin{table}[!h]
    \centering
    \begin{tabular}{|l|l|}
    \hline
        Hyperparameter & Value \\
    \hline
        Effective batch size & 8 \\
        \texttt{per\_device\_train\_batch\_size} & 2 \\
        \texttt{gradient\_accumulation\_steps} & 1 \\
        \texttt{learning\_rate} & \texttt{1.0e-6} \\
        \texttt{lr\_scheduler\_type} & \texttt{cosine} \\
        \texttt{warmup\_ratio} & 0.1 \\
        \texttt{bf16} & \texttt{true} \\
        \texttt{max\_steps} & 500 \\
    \hline
    \end{tabular}
    \caption{Hyperparameter settings for SFT.}
    \label{tab:appendix:sft_hyperpara}
\end{table}

\mypara{RL training.}
For the \texttt{QwenVL} model, we perform RL training using the \texttt{verl} package. The hyperparameter settings are reported in Table~\ref{tab:appendix:qwenvl_rl_hyperpara}. Each experiment is trained on 4 $\times$ 80GB A100 GPUs, with \texttt{ppo\_micro\_batch\_size\_per\_gpu} set to 10 and \texttt{log\_prob\_micro\_batch\_size\_per\_gpu} set to 80. Each \texttt{GC} run takes approximately 30 hours, while each \texttt{Sudoku} run takes approximately 50 hours, mainly because \texttt{Sudoku} examples have longer output sequences.

\begin{table}[!h]
    \centering
    \begin{tabular}{|l|l|}
    \hline
        Hyperparameter & Value \\
    \hline
        \texttt{lr} & \texttt{1e-6} \\
        \texttt{train\_batch\_size} & 128 \\
        \texttt{ppo\_mini\_batch\_size} & 128 \\
        \texttt{use\_kl\_loss} & \texttt{true} \\
        \texttt{kl\_loss\_coef} & 0.01 \\
        \texttt{total\_training\_steps} & 100 \\
    \hline
    \end{tabular}
    \caption{Hyperparameter settings for QwenVL RL.}
    \label{tab:appendix:qwenvl_rl_hyperpara}
\end{table}

For the \texttt{InternVL} model, due to compatibility issues, we use the \texttt{verl-internvl} implementation from \texttt{verl-internvl} \citep{verl_internvl_github} and largely follow its default hyperparameter settings to ensure stable performance. The hyperparameter settings are reported in Table~\ref{tab:appendix:internvl_rl_hyperpara}. We also train on 4 $\times$ 80GB A100 GPUs, with \texttt{use\_dynamic\_bsz} set to \texttt{true}, \texttt{ppo\_max\_token\_len\_per\_gpu} set to 20480, and \texttt{log\_prob\_micro\_batch\_size\_per\_gpu} set to 16. Training is faster than QwenVL RL: each \texttt{GC} run takes approximately 18 hours, while each \texttt{Sudoku} run takes approximately 30 hours.

\begin{table}[!h]
    \centering
    \begin{tabular}{|l|l|}
    \hline
        Hyperparameter & Value \\
    \hline
        \texttt{lr} & \texttt{1e-6} \\
        \texttt{ppo\_mini\_batch\_size} & 128 \\
        \texttt{ppo\_micro\_batch\_size\_per\_gpu} & 16 \\
        \texttt{use\_kl\_loss} & \texttt{false} \\
        \texttt{loss\_mode} & \texttt{gspo} \\
        \texttt{rollout.n} & 5 \\
        \texttt{total\_training\_steps} & 100 \\
    \hline
    \end{tabular}
    \caption{Hyperparameter settings for InternVL RL.}
    \label{tab:appendix:internvl_rl_hyperpara}
\end{table}

\mypara{Inference and evaluation.}
For all experiments, we run inference using the \texttt{vllm} package with \texttt{max\_tokens}=10000, \texttt{temperature}=0.7, \texttt{top\_p}=1.0, \texttt{top\_k}=-1. Standard inference and counterfactual inference for measuring reasoning accuracy each take approximately 30 minutes on a single A100 GPU. Evaluation is performed using rule-based scripts based on ground-truth perception annotations.
For all experiments, we report a single run due to the high computational cost. However, since each task uses a large controlled test set ($N=500$), we expect the reported results to provide reliable estimates of model performance.

\begin{figure*}[!t]
    \centering
    \includegraphics[width=\textwidth]{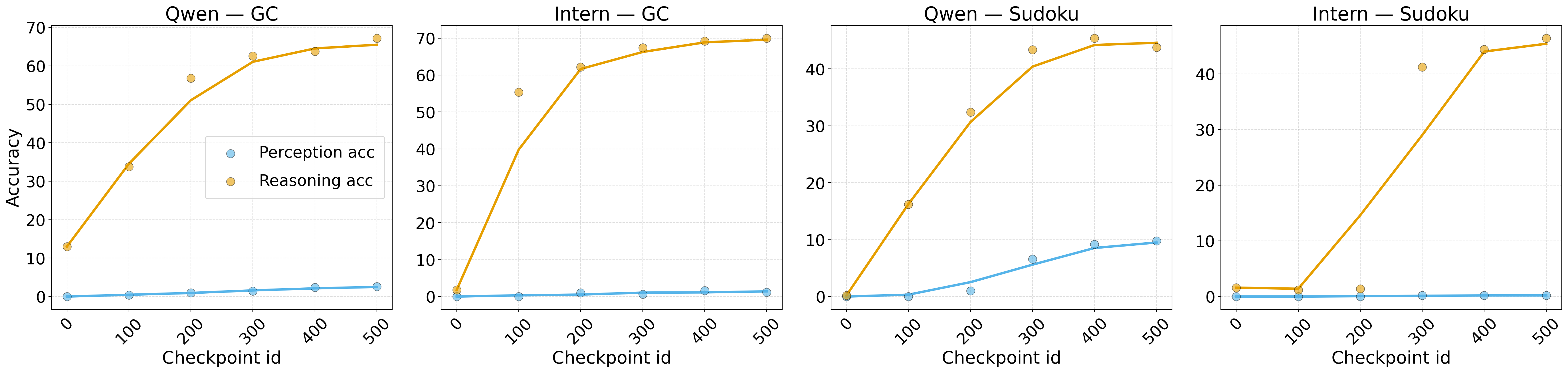}
    \caption{SFT optimization curves across all four model-task settings. This figure extends Figure~\ref{fig:sft_gap} in the main text.}
    \label{fig:appendix:sft_optimization_curves}
\end{figure*}

\begin{figure*}[!t]
    \centering
    \includegraphics[width=\textwidth]{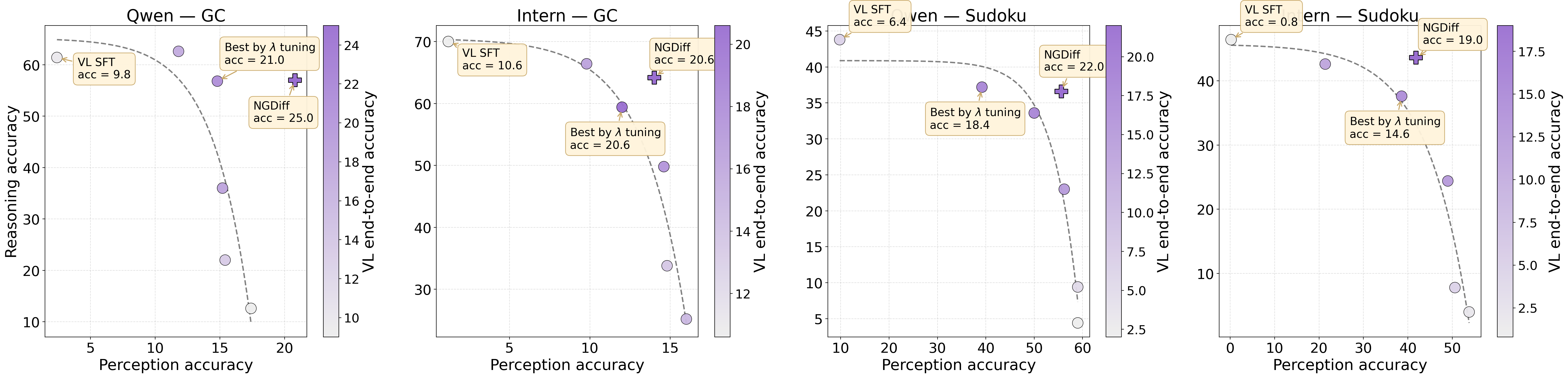}
    \caption{Perception-reasoning trade-off results for SFT loss reweighting across all four model-task settings. This figure extends Figure~\ref{fig:sft_tradeoff} in the main text.}
    \label{fig:appendix:sft_tradeoff}
\end{figure*}

\begin{figure*}[!t]
    \centering
    \includegraphics[width=\textwidth]{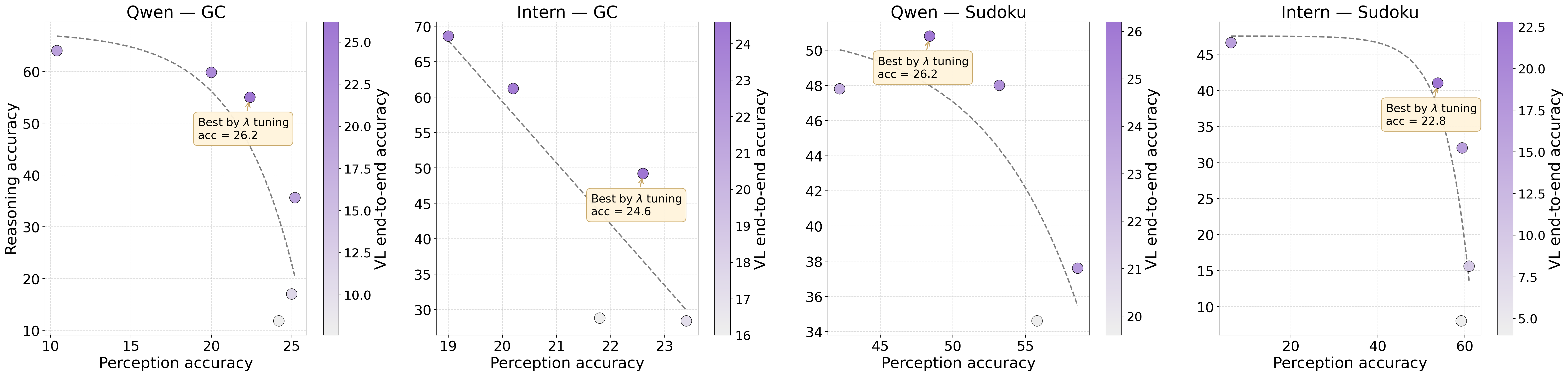}
    \caption{Perception-reasoning trade-off results after stronger perception initialization across all four model-task settings. }
    \label{fig:appendix:sft_tradeoff_strong_init}
\end{figure*}

\section{Additional Results}

\subsection{Full Results}
\label{sec:appendix:full_results}

\begin{figure*}[!t]
    \centering
    \includegraphics[width=\textwidth]{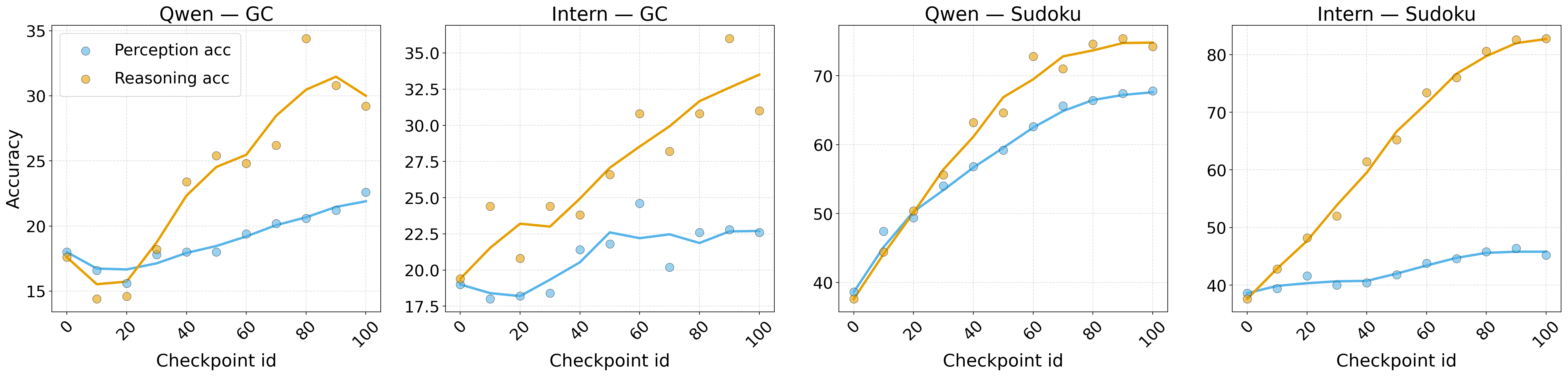}
    \caption{RL optimization curves across all four model-task settings. This figure extends Figure~\ref{fig:rl_gap} in the main text.}
    \label{fig:appendix:rl_optimization_curves}
\end{figure*}

\begin{figure*}[!t]
    \centering
    \includegraphics[width=\textwidth]{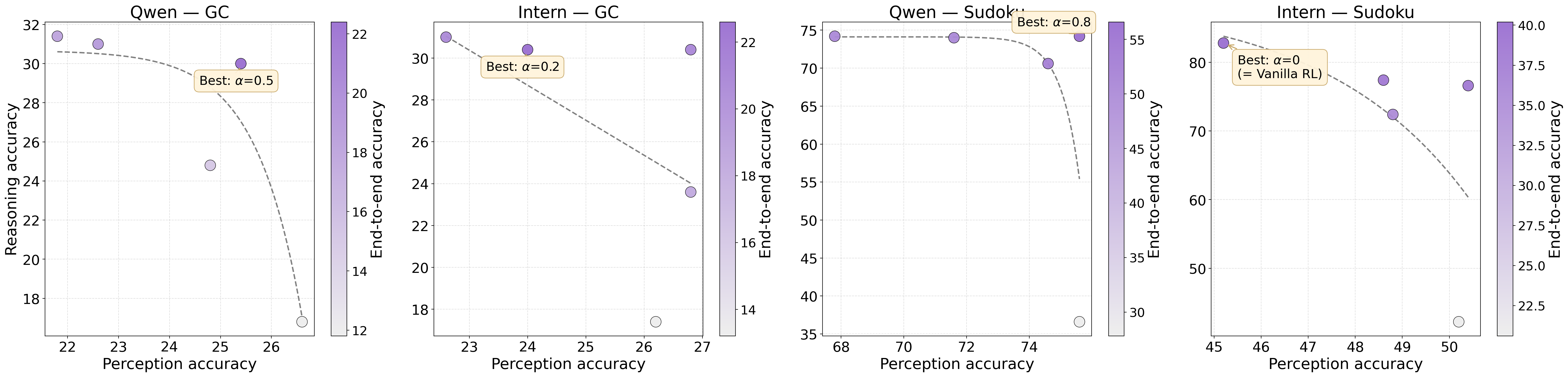}
    \caption{Full perception-reasoning trade-off results for RL reward augmentation across all four model-task settings. This figure extends Figure~\ref{fig:rl_tradeoff} in the main text.}
    \label{fig:appendix:rl_tradeoff}
\end{figure*}

In the main paper, we present representative results for selected model-task settings. Here, we provide the corresponding results across all four settings: \texttt{QwenVL-GC}, \texttt{InternVL-GC}, \texttt{QwenVL-Sudoku}, and \texttt{InternVL-Sudoku}. Specifically, Figure~\ref{fig:appendix:sft_optimization_curves} extends the SFT optimization curves in Figure~\ref{fig:sft_gap}, Figure~\ref{fig:appendix:sft_tradeoff} extends the SFT loss-reweighting trade-off in Figure~\ref{fig:sft_tradeoff}, Figure~\ref{fig:appendix:rl_optimization_curves} extends the RL optimization curves in Figure~\ref{fig:rl_gap}, and Figure~\ref{fig:appendix:rl_tradeoff} extends the RL reward-augmentation trade-off in Figure~\ref{fig:rl_tradeoff}. To account for run-to-run variance in RL, we repeat the RL experiment with perception-aware reward augmentation ($\alpha=0.8$) on the Sudoku task five times, achieving a final accuracy of $54.7 \pm 2.0$ (mean $\pm$ std), with the variation remaining smaller than the performance gain of up to 6 percentage points.

\mypara{Robustness to stronger perception initialization.} For the SFT results, one concern is that the trade-off emerges only from weak initial perception, limiting its relevance to practical post-training. We thus add a perception-only SFT stage before loss reweighting to provide a stronger perception initialization. As shown in Figure~\ref{fig:appendix:sft_tradeoff_strong_init}, a similar perception-reasoning trade-off persists with end-to-end performance peaking at an intermediate perception weight. This suggests that the trade-off stems from joint perception-reasoning optimization rather than an artifact of weak initialization.

\subsection{Token Reweighting for RL}
\label{sec:appendix:token_reweighting_for_rl}

In the main text, we show that token imbalance explains the perception-reasoning asymmetry in SFT, but not in RL. Here, we describe the RL token-reweighting intervention in more detail and provide the full optimization trajectory.

Recall that standard GRPO optimizes the token-averaged objective
\begin{align*}
\mathcal{J}_\text{GRPO}
=
\mathbb{E}\left[
\frac{1}{G}
\sum_{i=1}^{G}
\frac{1}{|\mathbf{y}^{(i)}|}
\sum_{t=1}^{|\mathbf{y}^{(i)}|}
\rho_{i,t}(\theta)\hat{A}_{i,t}
\right],
\end{align*}
where each response is decomposed into perception tokens $\mathbf{p}^{(i)}$ and reasoning tokens $\mathbf{r}^{(i)}$, i.e., $\mathbf{y}^{(i)}=[\mathbf{p}^{(i)},\mathbf{r}^{(i)}]$. To mirror the SFT loss-reweighting intervention, we first decompose the GRPO objective into perception and reasoning components:
\begin{align*}
\mathcal{J}_p
&=
\mathbb{E}\left[
\frac{1}{G}
\sum_{i=1}^{G}
\frac{1}{|\mathbf{p}^{(i)}|}
\sum_{y_t \in \mathbf{p}^{(i)}}
\rho_{i,t}(\theta)\hat{A}_{i,t}
\right], \\
\mathcal{J}_r
&=
\mathbb{E}\left[
\frac{1}{G}
\sum_{i=1}^{G}
\frac{1}{|\mathbf{r}^{(i)}|}
\sum_{y_t \in \mathbf{r}^{(i)}}
\rho_{i,t}(\theta)\hat{A}_{i,t}
\right].
\end{align*}
We then define the reweighted GRPO objective as
\begin{align*}
\mathcal{J}_{\text{GRPO},\lambda}
=
\lambda \mathcal{J}_p + (1-\lambda)\mathcal{J}_r,
\end{align*}
where $\lambda \in [0,1]$ controls the relative emphasis on perception-token updates. %

In practice, we implement this objective by keeping the original token-averaged GRPO form and rescaling the token-level advantages. Specifically, we define
\begin{align*}
\hat{A}^{(\lambda)}_{i,t}
=
\begin{cases}
\lambda \cdot \dfrac{|\mathbf{y}^{(i)}|}{|\mathbf{p}^{(i)}|}\hat{A}_{i,t},
& y_t \in \mathbf{p}^{(i)}, \\[1.2em]
(1-\lambda) \cdot \dfrac{|\mathbf{y}^{(i)}|}{|\mathbf{r}^{(i)}|}\hat{A}_{i,t},
& y_t \in \mathbf{r}^{(i)}.
\end{cases}
\end{align*}
Substituting $\hat{A}^{(\lambda)}_{i,t}$ into the standard token-averaged objective,
\begin{align*}
\mathcal{J}_{\text{GRPO},\lambda}
=
\mathbb{E}\left[
\frac{1}{G}
\sum_{i=1}^{G}
\frac{1}{|\mathbf{y}^{(i)}|}
\sum_{t=1}^{|\mathbf{y}^{(i)}|}
\rho_{i,t}(\theta)\hat{A}^{(\lambda)}_{i,t}
\right],
\end{align*}
recovers the decomposed reweighted objective above. Thus, although the intervention is conceptually an objective-level reweighting between perception and reasoning terms, our implementation applies it through advantage rescaling.

\begin{figure}[!t]
    \centering
    \includegraphics[width=\linewidth]{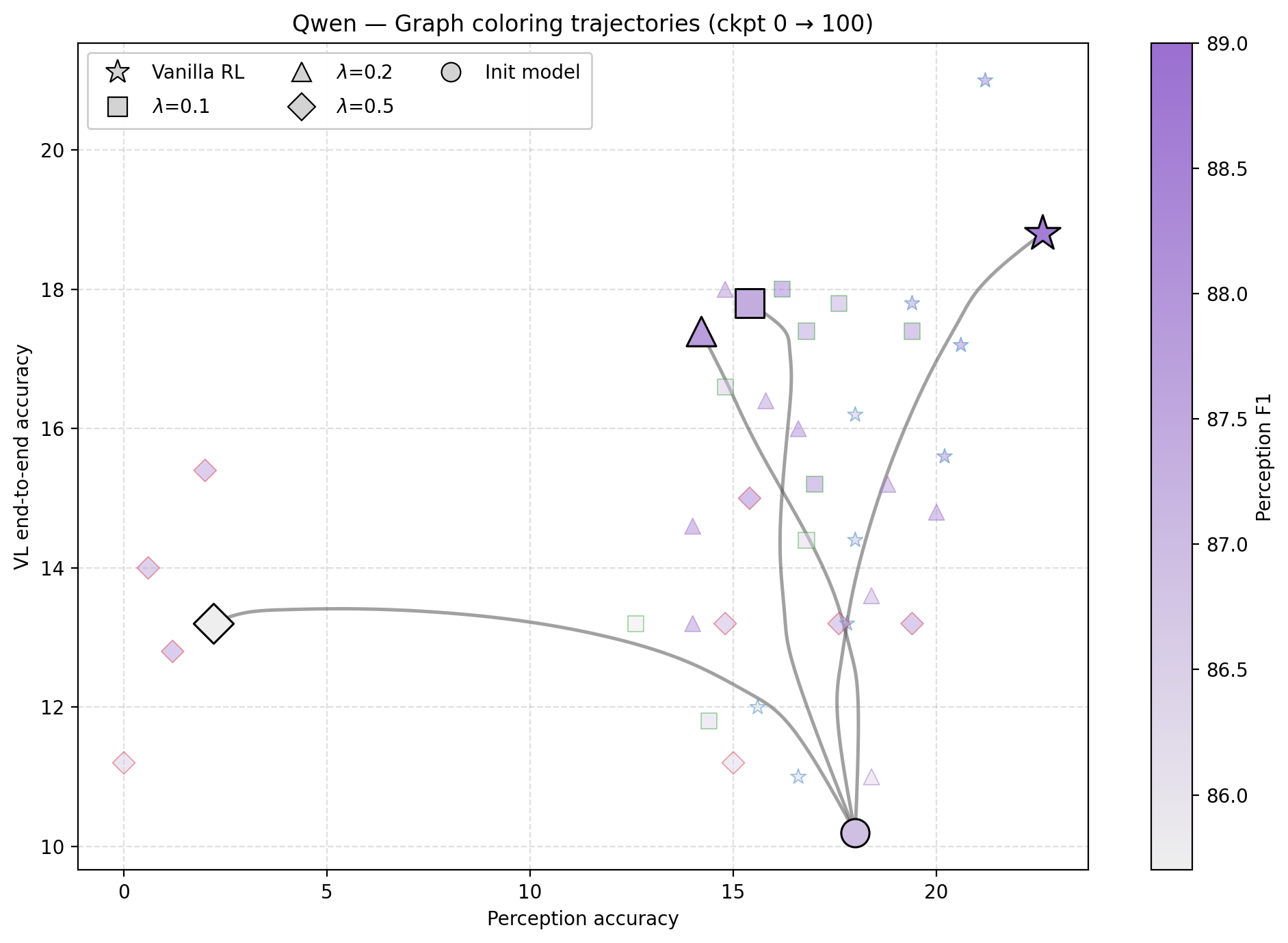}
    \caption{Optimization trajectories of RL token reweighting. Unlike SFT loss reweighting, directly increasing the weight on perception-token policy-gradient terms does not improve perception or end-to-end accuracy, and instead degrades training. This indicates that RL asymmetry is not primarily driven by token imbalance.}
    \label{fig:appendix:token_reweighting_in_rl}
\end{figure}

Figure~\ref{fig:appendix:token_reweighting_in_rl} shows the optimization trajectories under different values of $\lambda$. Unlike in SFT, token reweighting does not mitigate the perception-reasoning asymmetry in RL. Instead, increasing the perception-token weight destabilizes training and degrades both perception and end-to-end performance. This suggests that the RL asymmetry is not primarily caused by the number of perception tokens, even though perception remains a small fraction of the output sequence. Rather, the dominant issue is reward coupling: the outcome reward provides a weaker and noisier learning signal for perception than for reasoning.

\subsection{Surrogate Perception Rewards}
\label{sec:appendix:surrogate_rewards_impl}

\mypara{CLIP-based perception reward.}
For the CLIP-based surrogate, following \citet{chen2025perception}, we score the generated perception by its image-text similarity to the input image. Specifically, we encode the image and the extracted perception segment with CLIP and use their cosine similarity as the surrogate perception reward. This implementation is lightweight and does not require additional rollouts or teacher queries. However, because CLIP similarity is coarse-grained, it may fail to distinguish fine-grained structural errors in canonical perception representations, such as missing graph edges or incorrect Sudoku entries.

\mypara{Self-rewarding perception reward.}
Following \citet{li2025selfrewardingvisionlanguagemodelreasoning}, we construct a self-reward by testing whether the model's generated perception is sufficient for downstream reasoning. Specifically, for each generated CoT, we extract the perception segment and run a secondary rollout in which the model reasons only from this textual perception representation, without access to the original image. We then use the final-answer accuracy of this secondary rollout as the surrogate perception reward. The original self-rewarding setup uses a single rollout ($N=1$). We additionally evaluate a higher-budget variant with $N=4$ secondary rollouts per perception and use the average final-answer accuracy as the reward. This provides a more reliable surrogate signal, at the cost of additional inference budget. For training, we build on the official Vision-SR1 codebase.

\mypara{Teacher-based perception reward in synthetic settings.}
We also implement a teacher-based surrogate reward inspired by \citet{xiao2025perceptionr1advancingmultimodalreasoning}. Since our tasks use simplified canonical perception representations, we do not require a teacher-as-judge formulation. Instead, we directly query the teacher model, \texttt{gemini-3-flash-preview}, to generate a canonical perception representation from the image, and compute the reward by comparing the model's predicted perception against this teacher-generated representation. This yields a perception reward in the same format as our ground-truth perception reward, allowing us to use the standard RL training codebase without additional judge-model infrastructure.

\mypara{Teacher-based perception reward in realistic settings.} In realistic settings, however, a canonical perception representation is generally unavailable, so the comparison-based teacher reward described above no longer applies. For the experiments in \S\ref{sec:experiments:realworld}, we therefore adopt a judge-style formulation: we prompt the teacher model, \texttt{Qwen3-VL-30B-A3B-Instruct}, with the input image, the question, and the perception component extracted from the student's output, and ask it to judge whether the perception is both \textit{faithful} to the image and \textit{sufficient} for answering the question. The teacher returns a binary score, which we use directly as the surrogate perception reward in the augmented objective $R_\alpha$. The full judge prompt is shown below:

\newtcblisting{promptbox}{%
  listing only,
  breakable,
  enhanced,
  colback=gray!5,
  colframe=black,
  boxrule=0.5pt,
  arc=2pt,
  boxsep=0pt,
  left=3pt,
  right=3pt,
  top=3pt,
  bottom=3pt,
  listing options={
    basicstyle=\ttfamily\small,
    breaklines=true,
    breakatwhitespace=true,
    breakindent=0pt,
    breakautoindent=false,
    columns=fullflexible,
    keepspaces=true,
    xleftmargin=0pt,
    xrightmargin=0pt,
    aboveskip=0pt,
    belowskip=0pt,
    resetmargins=true,
    showstringspaces=false,
    moredelim={**[is][\color{orangehl}]{@|}{|@}}
  }
}

\begin{promptbox}
You are given an image, a question about the image, and a description of the image written by another AI model.
Decide whether the description contains enough ACCURATE visual information to correctly answer the question - i.e. is it SUFFICIENT and faithful for THIS specific question (no hallucinated details, and none of the details needed to answer the question are missing or wrong).
Answer with a single word: 'yes' if the description is sufficient and accurate for the question, or 'no' otherwise.

Question:
{question}

Description:
{caption}
\end{promptbox}

\end{document}